\pdfoutput=1

\documentclass[11pt]{article}

\usepackage{acl}

\usepackage{times}
\usepackage{latexsym}

\usepackage{amsmath}
\usepackage{graphicx}
\usepackage{enumitem}

\usepackage[T1]{fontenc}

\usepackage[utf8]{inputenc}

\usepackage{microtype}
\usepackage{booktabs}
\usepackage{tabularx}
\usepackage{float}
\usepackage{amsmath}
\usepackage{caption}
\usepackage{subcaption}
\usepackage{algorithm}
\usepackage{algpseudocode}

\usepackage{enumitem}
\usepackage[most]{tcolorbox}
\usepackage{graphicx}
\usepackage{amsfonts}
\usepackage{subcaption}
\usepackage{bm}
\usepackage{makecell}

\usepackage{inconsolata}
\usepackage{multirow,tabularx}

\usepackage{listings}
\usepackage{xcolor}

\definecolor{codegreen}{rgb}{0,0.6,0}
\definecolor{codegray}{rgb}{0.5,0.5,0.5}
\definecolor{codepurple}{rgb}{0.58,0,0.82}
\definecolor{backcolour}{rgb}{0.95,0.95,0.92}

\lstdefinestyle{mystyle}{
    backgroundcolor=\color{backcolour},   
    commentstyle=\color{codegreen},
    keywordstyle=\color{magenta},
    numberstyle=\tiny\color{codegray},
    stringstyle=\color{codepurple},
    basicstyle=\ttfamily\footnotesize,
    breakatwhitespace=false,         
    breaklines=true,                 
    captionpos=b,                    
    keepspaces=true,                 
    numbers=left,                    
    numbersep=5pt,                  
    showspaces=false,                
    showstringspaces=false,
    showtabs=false,                  
    tabsize=2
}

\title{A Graph Perspective to Probe Structural Patterns of \\Knowledge in Large Language Models}

\author{
    Utkarsh Sahu\textsuperscript{1}, Zhisheng Qi\textsuperscript{1}, Yongjia Lei\textsuperscript{1}, Ryan A. Rossi\textsuperscript{2}, Franck Dernoncourt\textsuperscript{2},\\ \bfseries Nesreen K. Ahmed\textsuperscript{3}, Mahantesh M Halappanavar\textsuperscript{4}, Yao Ma\textsuperscript{5}, Yu Wang\textsuperscript{1}\\
    \normalfont \textsuperscript{1}University of Oregon, \textsuperscript{2}Adobe Research, \textsuperscript{3}Cisco AI Research,\\ \textsuperscript{4}Pacific Northwest National Laboratory, \textsuperscript{5}Rensselaer Polytechnic Institute\\
    \texttt{\{utkarsh, charq, yongjia, yuwang\}@uoregon.edu, \{ryrossi, dernonco\}@adobe.com,} \\
  \texttt{nesahmed@cisco.com,  hala@pnnl.gov, may13@rpi.edu}
}

\begin{document}
\pagestyle{plain}
\maketitle
\thispagestyle{plain}
\begin{abstract}
Large language models have been extensively studied as neural knowledge bases for their knowledge access, editability, reasoning, and explainability. However, few works focus on the structural patterns of their knowledge. Motivated by this gap, we investigate these structural patterns from a graph perspective.
We quantify the knowledge of LLMs at both the triplet and entity levels, and analyze how it relates to graph structural properties such as node degree. Furthermore, we uncover the knowledge homophily, where topologically close entities exhibit similar levels of knowledgeability, which further motivates us to develop graph machine learning models to estimate entity knowledge based on its local neighbors.
This model further enables valuable knowledge checking by selecting triplets less known to LLMs. Empirical results show that using selected triplets for fine-tuning leads to superior performance. Our code is publicly available \href{https://github.com/utkarshxsahu/kgc}{here}.

\end{abstract}

\section{Introduction}
Large Language Models (LLMs) have emerged as powerful knowledge bases by encoding world knowledge within their neural parameters~\cite{kadavath2022language, pezeshkpour2023measuring, yin2023large}. This world knowledge allows LLMs to generate contextually relevant and factually rich responses to natural language prompts that serve real-world applications such as question-answering, fact-checking, and reasoning and planning. To more wisely leverage this capability, researchers have been probing LLMs' knowledge from various aspects~\cite{alkhamissi2022review, zheng2023kglens}, including consistency, editability, reasoning, and explainability. These probing efforts have inspired adaptive retrieval, knowledge checking and editing, confidence calibration, and hallucination detection~\cite{si2023knowledge, farquhar2024detecting, ahdritz2024distinguishing}.

\begin{figure}[t!]
    \centering
    \includegraphics[width=1\columnwidth]{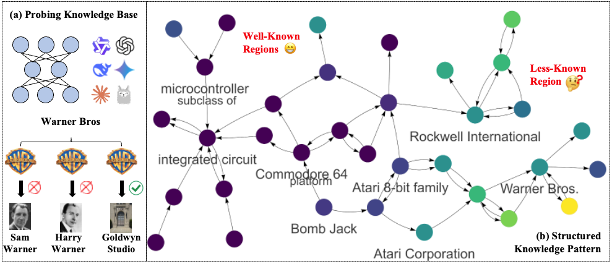}
    \caption{(a) Prompting LLMs to check their knowledge about each triplet and aggregate them to obtain entity knowledgeabilty; (b) These scores are assigned to graph nodes, enabling analysis of structural patterns such as knowledge imbalance (depicted in darker/lighter color), and knowledge homophly where topologically close entities possess similar levels of knowledgeability.}
    \vspace{-3ex}
    \label{fig-motivation}
\end{figure}

Despite the above progress~\cite{kadavath2022language, pezeshkpour2023measuring, zheng2024large}, few have examined structural patterns of LLMs' knowledge. Inspired by cognitive neuroscience~\cite{liu2025advances}, which has uncovered structured patterns in human knowledge organization, such as semantic networks that cluster related concepts~\cite{huth2016natural, hoedemaker2017onset}, specialized brain regions for specific information categories~\cite{binder2009semantic}, and topographic maps for sensory inputs~\cite{garvert2017map}, we hypothesize that similar structured patterns exist within LLMs. Probing these structural patterns provides critical insights into how knowledge is stored, retrieved, and reasoned in LLMs. For example, such understanding could support efficient knowledge retrieval and editing by leveraging structured knowledge organization.

Given the criticality of understanding the structural patterns of knowledge in LLMs and the limited exploration in this field, we take a fresh graph-based perspective to uncover the structural patterns of knowledge encoded in LLMs as shown in Figure~\ref{fig-motivation}. Building on these derived structural patterns, we develop graph machine learning models to identify more informative knowledge for fine-tuning LLMs. Our key contributions are as follows:
\begin{itemize}[leftmargin=*]
    \item \textbf{Novel Graph Perspective to Probe Structural Patterns of LLM Knowledge:} We introduce a novel graph-based approach to analyze structural patterns of knowledge in LLMs. Specifically, we define two knowledgeability metrics to quantify LLMs' knowledge at the triplet and entity levels.

    \item \textbf{Discovery of Novel Structural Patterns:} Several novel patterns are revealed, including entity knowledge imbalance, positive correlations between entity degree and knowledgeability, and knowledge homophily, where topologically proximate entities exhibit similar knowledgeability.

    \item \textbf{Graph Learning for Knowledge Prediction and Checking:} We design graph-based regression models to estimate LLM knowledgeability scores for each entity by leveraging its local neighborhood context. These predicted scores are then used to prioritize high-value triplet facts for more effective LLM fine-tuning.

\end{itemize}

\section{Method}
Given a graph $\mathcal{G} = (\mathcal{V}, \mathcal{R}, \mathcal{F})$ with $\mathcal{V}/\mathcal{R}/\mathcal{F}$ being the set of entities/relations/facts. Each fact is represented as a triplet $(v_i, r_{ij}, v_j)$ with $v_i/v_j \in \mathcal{V}$ being the head/tail entities, and $r_{ij} \in \mathcal{R}$ being their relation. We define the LLMs' knowledgeability for a given triplet $(v_i, r_{ij}, v_j)$/entity $v_i$ as $\mathcal{K}(v_i, r_{ij}, v_j)/\mathcal{K}(v_i)$, measuring the extent to which the LLM is aware of the triplet fact or entity. Regarding graph structural properties, the degree and clustering coefficient of an entity $v_i$ are denoted as $d_{v_i}$ and $c_{v_i}$. We define the neighbor entity set $\mathcal{N}(v_i)$ of $v_i$ as the set of entities directly connected to $v_i$ and the neighbor triplet set $\mathcal{T}(v_i)$ of $v_i$ as the set of triplets in which $v_i$ appears as either the head/tail entity. Next, we introduce knowledgeability measurement at the triplet/entity levels.

\subsection{Triplet Knowledgeability}\label{sec-triplet-KG}
Inspired by prior work~\cite{kadavath2022language, alkhamissi2022review, pezeshkpour2023measuring}, we transform each triplet $(v_i, r_{ij}, v_j)$ into a natural language statement and prompt LLMs to assess whether they recognize the fact. The response of LLMs is recorded as a binary value with \texttt{True}/\texttt{False} mapped to 1/0, indicating the knowledgeability of LLM about the triplet $\mathcal{K}_{(v_i, r_{ij}, v_j)}$.

To handle temporal triplets with time information $(v_i, r_{ij}, v_j, t)$ (e.g., ``Donald Trump made a visit to China on 2017-11-08.''), we extend the prompt to explicitly incorporate timestamps, allowing us to consider the temporal impact on LLM knowledgeability. The template of the initial prompt is shown as below with its temporal variation attached in Appendix~\ref{app-temporal-prompt}:

\begin{tcolorbox}[
colback=blue!10!white, 
colframe=blue!80!black, 
title=Prompt 1: LLM-based Triplet Evaluation, 
boxsep=0.75mm, 
left=0.75mm, 
right=0.75mm, 
top=0.75mm, 
bottom=0.75mm, 
float=htbp!,     
floatplacement=tbp, 
]
\scriptsize
\textbf{System Message:} Evaluate the statement based on your knowledge and respond with \texttt{True} or \texttt{False}.\\[2pt]
\textbf{Given:} Triplet $\mathcal{T}=(\mathit{sub},\,\mathit{rel},\,\mathit{obj})$.\\[2pt]
\textbf{Relational Template Map:} $T:\,\mathit{rel}\mapsto\text{“\{\textit{sub}\} … \{\textit{obj}\}”}$.\\[2pt]
\textbf{Procedure:}
\begin{enumerate}[nosep,left=6pt]
  \item Retrieve relation-based template $t = T(\mathit{relation})$.
  \item Instantiate statement  
    $S = t[\{\mathit{sub}\}\!\to\!\mathit{sub},\;\{\mathit{obj}\}\!\to\!\mathit{obj}]$.
  \item Prompt \textbf{System Msg} + \textbf{User Msg:} $S$ to the LLM.
  \item Return “True” or “False.”
\end{enumerate}
\label{box:prompt_template}
\end{tcolorbox}




\subsection{Entity Knowledgeability and Homophily}
\begin{figure*}[t!]
    \centering
    \includegraphics[width=1\linewidth]{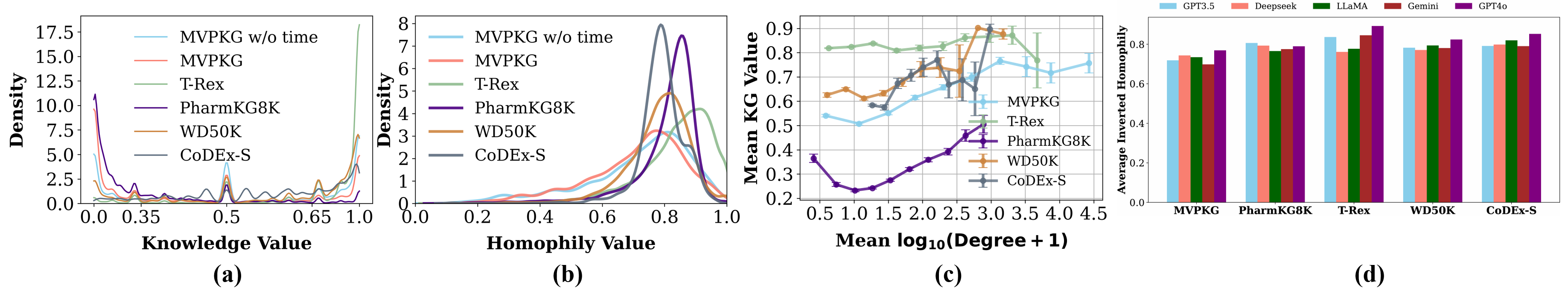}
    \vspace{-5.5ex}
    \caption{\textbf{(a)/(b)}: Distribution of node knowledgeability/homophily for each dataset; \textbf{(c)}: Node knowledgeability increases as node degree increases. The results here are based on GPT3.5, and results for other LLMs hold similar observations in Appendix~\ref{app-result}. \textbf{(d)}: Average homophily for all datasets given by different LLMs exceeds 0.6.}
    \label{fig-analysis}
    \vspace{-1ex}
\end{figure*}

Given the above triplet knowledgeability, we obtain the entity $v_i$'s knowledgeability score by aggregating the knowledgeability of all triplets in which $v_i$ is involved~\cite{jia2019triple, rings2022network}:
\begin{equation}
\small
    \mathcal{K}(v_i) = {\lvert \mathcal{T}(v_i)\rvert}^{-1}\sum_{(v_i, r_{ij}, v_j) \in \mathcal{T}(v_i)} \mathcal{K}(v_i,r_{ij},v_j)
\end{equation}
Note that the above neighborhood aggregation to obtain the knowledgeability score for each entity also applies to temporal triplets $(v_i, r_{ij}, v_j, t) \in \mathcal{T}(v_i)$, allowing us to account for the temporal impact when assessing an entity's knowledgeability. The change of knowledgeability after incorporating temporal information is shown in Figure~\ref{fig-analysis}(a).

Furthermore, we evaluate whether topologically close entities share similar knowledgeability, i.e., the homophily of entity knowledgeability $\mathcal{H}_{v_i}$. Inspired by existing homophily computation~\cite{wang2021tree, ma2021homophily}, we compute knowledgeability homophily as one minus the average absolute difference in knowledgeability between central node $v_i$ and its neighbors $\mathcal{N}(v_j)$ in the knowledge graph:
\begin{equation}
\small
    \mathcal{H}_{v_i} = 1 - \frac{1}{|\mathcal{N}(v_i)|}\sum_{v_j \in \mathcal{N}(v_i)}|\mathcal{K}(v_i) - \mathcal{K}(v_j)|,
\end{equation}

\subsection{Knowledgeability Regression with GNNs.}
Given the observed high homophily of entity knowledgeability scores in Figure~\ref{fig-analysis}(b), we further design GNN-based graph regression models to approximate the knowledgeability of unknown entities based on known ones. Specifically, given a fixed set of entities $\mathcal{V}^{\text{Train}}$ with known knowledgeability, our goal is to train a GNN model to estimate the entity knowledgeability with unknown scores. We perform message-passing ($\text{MP}$) and feature transformation ($\text{TR}$) followed by regression:
\begin{equation}\label{eq-propagation}
\small
    \widehat{\mathcal{K}}_i^l = \text{MP}^{l}(\{\widetilde{\mathcal{K}}_j^{l - 1}|v_j \in \mathcal{N}(v_i) \cup \{v_i\}\}), \widetilde{\mathcal{K}}_i^l = \text{TR}^{l}(\widehat{\mathcal{K}}_i^l),
\end{equation}
\begin{equation}\label{eq-loss}
\small
    \mathcal{L} = \frac{1}{|\mathcal{V}^{\text{Train}}|}\sum_{v_i \in \mathcal{V}^{\text{Train}}} \left\| \widetilde{\mathcal{K}}_i^l - \mathcal{K}_i \right\|_2^2,
\end{equation}
The initial node feature matrix is defined as $\widetilde{\mathcal{K}}^0 = [\mathcal{X}(v_1), \dots, \mathcal{X}(v_{|\mathcal{V}|})]^\top$, where each node feature $\mathcal{X}(v_i)$ is either a one-hot encoding or a dense text embedding obtained from pretrained language models. By training a regression model on a subset of entities $\mathcal{V}^{\text{Train}}$, we manage to estimate the knowledgeability of all entities without the need for resource and time-intensive knowledge probing via prompting LLMs across the entire entity set.




\section{Experiment}\label{sec-expr}
In this section, we quantify triplet/entity knowledgeability, analyze its correlation with structural properties of the underlying graphs, estimate knowledgeability using GNNs, and explore active selection strategies to identify high-valuable triplets for fine-tuning. We evaluate five representative LLMs: commercial ones such as GPT-3.5, 4o, Gemini-2.5 Flash, and two open-source models, LLaMA3.3-70B and DeepSeek-V3. These models are assessed across five knowledge graphs: MVPKG~\cite{mou2024unifying}, T-Rex~\cite{elsahar2018t}, PharmKG8K~\cite{zheng2021pharmkg}, WD50K~\cite{galkin2020message}, and CoDEx-S~\cite{safavi2020codex}. Among them, T-Rex, WD50K, and CoDEx-S represent general factual Wikipedia knowledge, whereas PharmKG8K and MVPKG focus on specialized pharmaceutical and political science. Further details on datasets and experimental configurations are in Appendix~\ref{app-setting}. We now present our key experimental findings. 

\textbf{Finding 1 - }Figure~\ref{fig-analysis}(a) presents the distribution of entity knowledgeability scores across various datasets. The scores exhibit a trimodal pattern with peaks at 0.0, 0.5, 1.0, corresponding to cases where none, some, or all of an entity’s triplets are recognized. 
%
These patterns exhibit clear domain-specific variation. Specialized datasets such as PharmKG8K and MVPKG are left-skewed, with a dominant peak at 0.0 reflecting LLM’s limited knowledge coverage in domains like pharmaceuticals and political science. In contrast, general-purpose datasets like T-Rex and WD50K are right-skewed, with most entities scoring 1.0, indicating substantial knowledge coverage in Wikipedia-based knowledge.
Comparing MVPKG and its temporal variant, MVPKG w/o time, we observe an increase in the proportion of entities with zero knowledgeability and a decrease in those scoring 1.0. This indicates challenges of LLMs in understanding time-sensitive knowledge~\cite{yuan2024back}.

\textbf{Finding 2 -} Figure~\ref{fig-analysis}(b)/(d) presents the node homophily distribution and the average graph homophily across several knowledge graphs. In Figure~\ref{fig-analysis}(b), these distributions are all right-skewed, with a peak around 0.8, suggesting that nodes and their neighbors tend to share similar knowledgeability scores. This high homophily property has enhanced graph machine learning in node-level prediction, such as node classification, and inspires our regression to predict entities' knowledge scores in Finding 3.
Furthermore, incorporating temporal information into MVPKG results in a slight shift to the left, indicating decreased neighbor score similarity. This shift indicates that the temporal dimension introduces greater complexity and finer knowledgeability distinctions between the nodes and their neighbors.
In addition, we compute the average graph homophily by averaging across all nodes and find that it consistently remains above 0.5 across different datasets and LLMs. This exhibits a general tendency for entities to be connected to others with similar knowledgeability scores. 
This finding reinforces the notion that the LLM’s factual recognition is not randomly distributed in the graph but is instead correlated among connected entities. 


\textbf{Finding 3 - }Figure~\ref{fig-analysis}(c) illustrates the relation between entity degree and knowledgeability. We observe a clear positive correlation, indicating that entities with higher degrees tend to exhibit greater knowledgeability in LLMs. This trend likely arises because high-degree entities are associated with more factual content and appear more frequently in pre-training corpora, increasing their likelihood of being learned during the LLM pre-training process. This observation aligns with findings showing accuracy disparities between popular and less popular entities~\cite{sun-etal-2024-head}. Notably, on the T-Rex dataset, the positive relationship remains but is much less pronounced. This is likely because T-Rex exclusively contains Wikipedia entities, which are generally well represented in LLM training corpora, even for less popular or low-degree entities.

\begin{table}[t!]
\small
\setlength\tabcolsep{1.5pt}
\centering
\caption{Regression of predicting node knowledgeability calculated by (1 - Mean Absolute Error between ground-truth and estimated knowledgeability scores). N/T-X represents the model X with input features being one-hot encoding (N)/textual embedding (T). The best performance is \textbf{bolded} and the second best is \underline{underlined}.}
\vspace{-2ex}
\begin{tabular}{l|c|c|c|c|c}
\toprule
\textbf{Model}
& \textbf{T-Rex} 
& \textbf{WD50K} 
& \textbf{Pharm} 
& \textbf{MVPKG(w/o t)} 
& \textbf{CoDEx} \\
\midrule
N-MLP  & 81\% & 78\% & 82\% & 72\% (70\%) & 84\% \\
N-GCN  & \underline{84\%} & \textbf{82\%} & \textbf{84\%} & 76\% (76\%) & 87\% \\
N-SAGE & 84\% & \underline{82\%} & \underline{84\%} & 76\% (77\%) & \textbf{87\%}
\\
\midrule
T-MLP  & 83\% & 78\% & 83\% & 76\% (77\%) & 86\% \\
T-GCN  & 84\% & 81\% & 84\% & \underline{78\%} (\textbf{80\%}) & \underline{87\%} \\
T-SAGE & \textbf{84\%} & 81\% & 84\% & \textbf{78\%} (\underline{79\%}) & 87\% \\
\bottomrule
\end{tabular}
\label{tab-regression}
\vspace{-2ex}
\end{table}

\begin{figure}[t!]
    \centering
    \includegraphics[width=1\linewidth]{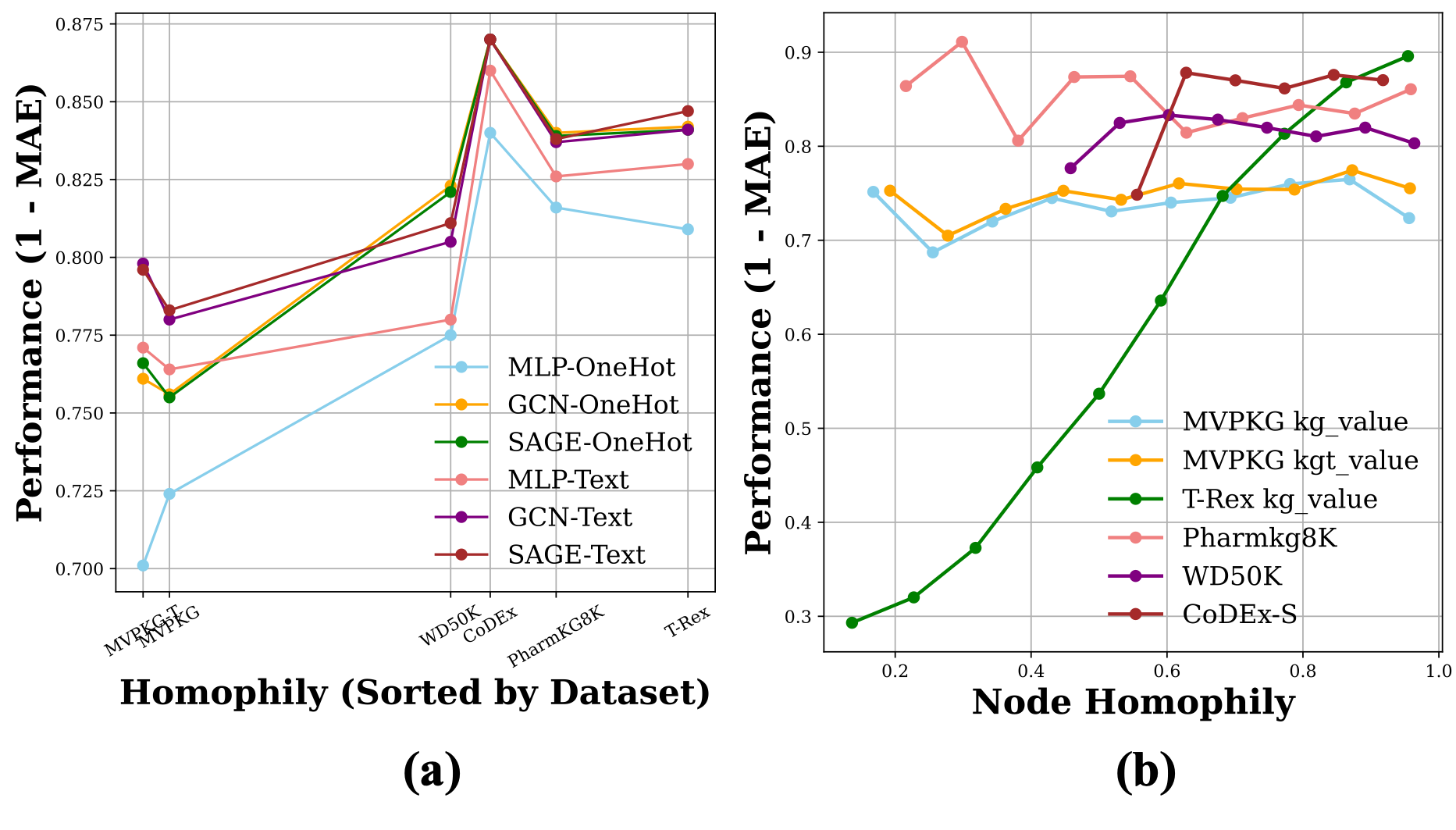}
    \vspace{-5ex}
    \caption{Relation between regression performance and homophily at \textbf{(a)} graph and \textbf{(b)} node level.}
    \label{fig-homophily-performance}
    \vspace{-3ex}
\end{figure}

\textbf{Finding 4 - }Table~\ref{tab-regression} demonstrates strong regression in predicting node knowledgeability, with absolute errors between 0.15 and 0.25. Comparing models using textual embeddings versus one-hot encodings reveals no consistent performance advantage, indicating that textual similarity between entities does not reliably reflect similarity in knowledgeability. In contrast, GNN-based models consistently outperform their MLP-based counterparts, underscoring the importance of incorporating neighborhood context for knowledgeability prediction. This result aligns with previous findings on the benefits of homophily in relational learning~\cite{ma2021homophily}. Figure~\ref{fig-homophily-performance}(a) visualizes a positive correlation between average regression performance and global graph homophily. However, in Figure~\ref{fig-homophily-performance}(b), while this trend holds for T-Rex, WD50K, and CoDEx-S, it is less apparent for PharmKG and MVPKG, suggesting that the effect of homophily may be dataset-dependent.


\begin{table}[t!]
\small
\setlength\tabcolsep{3.5pt}
\centering
\caption{Performance comparison between fine-tuning with random triplet selection (Random-FT) and with knowledgeability-based selection (Graph-FT), where triplets are ranked from high to low based on estimated knowledgeability. The best performance is \textbf{bolded} and the second best is \underline{underlined}.}
\vspace{-2ex}
\begin{tabular}{ll|ccc}
\toprule
\textbf{Dataset} & \textbf{Model} & \textbf{Base} & \textbf{Random-FT} & \textbf{Graph-FT} \\
\midrule
\multirow{3}{*}{\textbf{T-Rex}} & Llama3 8B & 63.25 & \underline{86.40} & \textbf{89.05} \\
 & Mistral 7B & 63.95 & \underline{81.85} & \textbf{91.90} \\
 & Qwen2.5 7B & 56.05 & \textbf{84.80} & \underline{83.25} \\
 \midrule
\multirow{3}{*}{\textbf{Pharm}} & Llama3 8B & 17.80 & \underline{34.85} & \textbf{36.95} \\
 & Mistral 7B & \underline{55.30} & 41.30 & \textbf{60.70} \\
 & Qwen2.5 7B & 39.50 & \underline{70.20} & \textbf{74.40}\\
  \midrule
\multirow{3}{*}{\textbf{WD50}} & Llama3 8B & 54.75 & \underline{57.75} & \textbf{58.75} \\
 & Mistral 7B & 42.87 & \textbf{56.25} & \underline{55.12 }\\
 & Qwen2.5 7B & 49.37 & \underline{63.00} & \textbf{64.75}\\
   \midrule
\multirow{3}{*}{\textbf{\makecell[c]{MVPKG \\ w/o t}}} & Llama3 8B & 26.10  & \underline{30.70}  & \textbf{44.50} \\
 & Mistral 7B & 52.30  & \underline{65.10} & \textbf{76.70}\\
 & Qwen2.5 7B & 37.60 & \underline{41.30} & \textbf{65.10}\\
    \midrule
 \multirow{3}{*}{\textbf{CoDEx}} & Llama3 8B & 64.87  & \textbf{78.75}  & \underline{75.62} \\
 & Mistral 7B & 58.50  & \underline{72.12} & \textbf{88.00} \\
 & Qwen2.5 7B & 62.37 & \underline{67.00} & \textbf{70.87} \\
 \midrule
 \multicolumn{2}{c|}{\textbf{Average Performance}} & 49.64 & \underline{62.09} & \textbf{69.04} \\
 \bottomrule
\end{tabular}
\label{tab-fine-tuning}
\vspace{-3ex}
\end{table}

We demonstrate a practical application of GNN-predicted knowledgeability scores to guide the selection of informative triplets for effective LLM fine-tuning. Specifically, we fine-tune three LLMs, LLaMA 3 8B, Mistral 7B, and Qwen 2.5 7B, across five datasets using two triplet selection strategies: Random-FT and Graph-FT. Both start by selecting the same initial 20\% of triplets for knowledge probing. Random-FT then randomly selects the remaining 80\%, while Graph-FT trains a GNN on the initial 20\% to estimate entity-level knowledgeability and selects additional triplets involving entities predicted to be less known (i.e., with lower knowledgeability scores). All experiments use identical hyperparameters within each dataset.

In Table~\ref{tab-fine-tuning}, both Random-FT and Graph-FT outperform the base models across all datasets. Notably, graph-FT consistently outperforms random-FT, underscoring the benefit of checking triplets with which the model is less familiar rather than redundantly reinforcing known knowledge. 



\section{Conclusion}
\vspace{-1ex}
This work introduces a novel graph-centric perspective by quantifying LLM knowledge at the triplet/entity levels and examining its relationship with graph structural properties. We uncover key insights, including a strong correlation between node degree and knowledgeability, and a high homophily degree, where topologically close nodes exhibit similar knowledgeability. These observations motivate the design of a graph machine learning model utilizing neighborhood information to predict entity-level knowledgeability. The predicted scores are then used to actively select more informative triplets for effective fine-tuning LLMs. 

\section{Limitations}
The limitations of this paper are as follows:

\begin{itemize}[leftmargin=*]
    \item \textbf{More applications}: The derived structural patterns are used solely to guide triplet selection for fine-tuning. However, these patterns hold broader potential. For instance, they could inform knowledge graph retrieval by identifying poor knowledge regions and prioritizing retrieving triplets there. Furthermore, this technique can also efficiently identify knowledge deficiency through structural correlations~\cite{song2025discovering}.

    \item \textbf{Limited to knowledge graphs}: The derived structural patterns currently apply only to knowledge graphs with explicitly defined entities and relations. However, real-world networks, such as social or citation networks, are often more complex and rich in textual information. Extending the entity/triplet-level knowledgeability estimation to these text-attributed graphs~\cite{wu2024stark} would broaden real-world applications.
\end{itemize}

\bibliography{reference}

\appendix

\newpage
\section{Appendix}\label{app}
In the appendix, we provide a comprehensive review of related work, covering three main areas: (1) Large Language Models (LLMs) as Knowledge Bases, (2) Knowledge Verification, and (3) the Topological Understanding of LLMs as Knowledge Bases. We then present detailed dataset statistics for all knowledge graphs used in this study. We describe the experimental setup employed to evaluate entity- and triplet-level knowledgeability. Finally, we include an in-depth analysis of different LLMs across various knowledge graphs, examining their knowledgeability distribution, homophily patterns, and correlations with structural graph properties.
\section{Related Work}
\subsection{LLM as Knowledge Base (KB)}
\cite{petroni2019language} was among the first works to propose that pretrained LMs encode factual knowledge retrievable via cloze prompts. Subsequent work such as ~\cite{roberts-etal-2020-much} fine-tuned LLM for closed-book QA to match external knowledge systems, \cite{heinzerling2020language} supported LMs as KBs by examining entity representations and paraphrase robustness, and~\cite{he2024can} demonstrated that LLMs trained on large-scale data could flexibly retrieve information, further bolstering the concept of LLMs as knowledge bases. This motivates the research on checking knowledge of LLMs as follows.


\begin{table*}[t!]
\centering
\caption{Statistics of the original knowledge graph and the sampled largest connected component.}
\setlength\tabcolsep{3.2pt}
\begin{tabular}{l|cc|cc|cc|cc}
\hline
\multirow{2}{*}{\textbf{Dataset}} & \multicolumn{2}{c|}{\# \textbf{Nodes}} & \multicolumn{2}{c|}{\# \textbf{Triplets}} & \multicolumn{2}{c|}{\# \textbf{Avg. Deg}} & \multicolumn{2}{c}{\# \textbf{Avg. CC}} \\
 & \textbf{Original} & \textbf{Sampled} & \textbf{Original} & \textbf{Sampled} & \textbf{Original} & \textbf{Sampled} & \textbf{Original} & \textbf{Sampled} \\ \hline
\textbf{T-Rex} & 3153568 & 46891 & 6566790 & 193781 & 4.16 & 8.26 & 0.1473 & 0.5170 \\
\textbf{WD50K} & 41334 & 5140 & 233838 & 34208 & 11.31 & 13.31 & 0.0996 & 0.1332 \\
\textbf{PharmKG8K} & 7262 & 6877 & 479902 & 98537 & 132.16 & 28.65 & 0.2512 & 0.0824 \\
\textbf{MVPKG} & 137117 & 9055 & 1857410 & 255697 & 12.46 & 28.24 & 0.0013 & 0.0140 \\
\textbf{MVPKG w/o t} & 137117 & 9055 & 1857410 & 116127 & 12.46 & 12.82 & 0.0013 & 0.0140 \\
\textbf{CoDEx-S} & \multicolumn{2}{c|}{2034} & \multicolumn{2}{c|}{36543} & \multicolumn{2}{c|}{35.93} & \multicolumn{2}{c}{0.0952} \\ \hline
\end{tabular}
\end{table*}

\subsection{Knowledge Checking}
To further evaluate this paradigm of LLM as KB, various knowledge checking methods have been developed, such as, factuality testing with TruthfulQA benchmark ~\cite{lin2021truthfulqa}, consistency and reliability ~\cite{zheng2024large}, calibration with self-assessed P(True) and P(I Know)~\cite{kadavath2022language}, information‐theoretic probing using entropy and KL‐divergence~\cite{pezeshkpour2023measuring}, systematic KG-based evaluation via autogenerated QA from graphs ~\cite{luo2023systematic}, and evaluating factuality hallucinations by using false premise questions ~\cite{zhu2024kg}.
These approaches look into knowledge and trustworthiness checking but treat the model as a black box, leaving its underlying structural patterns unexplored.

\subsection{Topological Understanding of LLM-KB}
Some important initial work has looked into local structures of LLMs. ~\cite{geva2020transformer} presented that feed-forward layers act like key–value memories for specific facts. Then ~\cite{meng2022locating} presented that factual associations are often localized and editable within mid-layer feed-forward modules. ~\cite{dai2021knowledge} proposed that factual knowledge is stored in pretrained Transformers in form of knowledge neurons. ~\cite{mruthyunjaya2023rethinking} evaluated LLMs on structural indicators such as, symmetry, hierarchy and path among others and show that they often fail on relational tests. These studies demonstrate that some implicit structure exists and yet none characterizes the graph topology or structural patterns of an LLM’s knowledge base. 


\section{Dataset Statistics}\label{app-expr}
Our experiments are designed to evaluate and compare the knowledgeability of the LLM across multiple datasets. We illustrate our process on five datasets: \textbf{MVPKG} (covering U.S.\ legislative, election, diplomatic data, etc.), \textbf{T-Rex} (containing large-scale high-quality alignments between DBpedia abstracts and Wikidata triples), \textbf{PharmKG8K} (biomedical knowledge graph), \textbf{WD50K} (dataset derived from Wikidata statements), and \textbf{CoDEx-S} (extracted from Wikidata and Wikipedia).

\begin{itemize}[leftmargin=*]
    \item \textbf{MVPKG}~\cite{mou2024unifying}: The MVPKG dataset encompasses U.S. legislative, election, and diplomatic data as well as conceptual knowledge from Wikidata. It originally contains 1,857,410 triplets, 137,117 entities, and 602 relations. Due to scale considerations, we extract the largest strongly connected component, which comprises 255,697 triplets, 9,055 entities, and 602 relations. The MVPKG dataset had a temporal attribute and was evaluated with the temporal component included and excluded. For each triplet, two prompts are generated (with time and without time). Consequently, each entity in MVPKG is assigned two knowledgeability scores corresponding to the two prompt variants for further analysis of the effect of inclusion of temporal information. All other datasets have only one knowledgeability score due to lack of temporal attributes.

    \item \textbf{T-Rex}~\cite{elsahar2018t}: The T-Rex dataset is constructed from Wikipedia abstracts aligned with Wikidata entities in English. It contains 6,566,790 unique triplets; the largest connected component comprises 193,781 triplets, 46,891 entities, and 423 relations. 

    \item \textbf{PharmKG8K}~\cite{zheng2021pharmkg}: The PharmKG8K multi-relational, attributed biomedical KG, composed of around 500,000 individual interconnections between genes, drugs, and diseases, with 29 relation types over a vocabulary of around 8000 disambiguated entities. Given the scope of the dataset, we used a strongly connected component of 98,537 edges, 6,877 entities, and 29 relations.

    \item \textbf{WD50k}~\cite{galkin2020message}: The WD50K dataset was created using the Wikidata RDF dump of August 2019. It has 233,838 edges and 41,334 entities. Since being extracted from Wikidata, there were 14,858 triplets common between the WD50K dataset and the T-Rex largest connected component selected. These were removed to make sure that common triplets were not overshadowing the result comparison between these datasets. Following that, the largest strongly connected component was selected for experimental purposes. This LCC had 34,208 edges, 5,140 entities, and 193 relations.

    \item \textbf{CoDEx-S}~\cite{safavi2020codex}: CoDEx is a collection of knowledge graph completion datasets extracted from Wikidata/Wikipedia, comprising three subsets of varying sizes. We select CoDEx-S due to its high proportion of triplets involving the ``occupation" relation, which poses greater challenges for LLMs, since individuals may hold multiple occupations. CoDEx-S contains 36,543 triplets, 2,034 entities, and 42 relations.
\end{itemize}

\section{Experimental Setting}\label{app-setting}
We describe the experimental setup for (1) measuring the triplet and entity knowledgeability, (2) training GNNs to predict knowledgeability scores, and adaptively selecting informative triplets to fine-tune LLMs.

\subsection{Measuring Knowledgeability Score}
\begin{itemize}[leftmargin=*]
    \item \textbf{Prompt Generation:}  
    Each triplet is converted into a natural language prompt using predefined templates based on the relation type, following~\cite{petroni2019language}. These templates were first generated by GPT o-1 mini using the relation and a few of its triplet examples to provide context, and then evaluated to make sure the template made semantic sense. \cite{luo2023systematic} used GPT3.5 for generating natural language prompts for triplets, validating that LLMs like GPT3.5 can be used for template or prompt generation. For MVPKG, both time-specific and non-time-specific prompts are created.
    
    \item \textbf{LLM Evaluation:}  
    The prompts are fed to the LLM, and responses are recorded as binary values (1 for true, 0 for false). This step enables us to quantify the LLM’s internalized knowledge regarding each triplet in a way that's scalable.
        
    \item \textbf{Aggregation to Entity-Level Scores:}  
    For every entity, triplet-level scores are aggregated to form the entity-level knowledgeability metric. In MVPKG, separate aggregations are performed for the two prompt types, giving two knowledgeability values for each entity. 
\end{itemize}

\subsection{Fine-Tuning: Random VS Graph}
The goal of this experiment is to evaluate whether fine-tuning LLMs on entities for which the model has low prior knowledge results in greater performance improvements than fine-tuning on randomly selected entities. We hypothesize that targeting entities about which the model knows less will produce a larger marginal improvement per example than fine-tuning on entities already well encoded in LLM's internal knowledge inherited during the pre-training phase.
\begin{itemize}[leftmargin=*]
    \item \textbf{Model and Evaluation Set:} To test this, we select three open source models: Llama 3.1 8B, Mistral v0.3 7B, and Qwen 2.5 7B, and constructed an evaluation set for each dataset by randomly sampling a fixed number of triplets. Each triplet is converted into a natural language prompt and is asked to LLM as a True/False evaluation task. Baseline performance is measured by querying each base model on this evaluation set prior to any fine-tuning. The performance metric is the percentage of correct responses by the model on the evaluation set.
    
    \item \textbf{Fine-Tuning Budget and Initial Query:} We then set a budget that the LLM can be fine-tuned on, and the size of this budget is adjusted according to the domain and size of the dataset. Twenty percent of the budget is reserved for an initial query set. To set up this initial set, we shuffle the entity list and iterate through it, adding all triples associated with the current entity until the 20\% quota is met and if an entity would overshoot the quota, we randomly subsample just enough of its triples to fill the gap.
    
    \item \textbf{Graph Fine-Tuning:} The triplets in this initial query set are posed to the base model, allowing us to calculate an entity-level knowledgeability score for the selected entities in the initial query.
    These entity scores are used to train a GraphSAGE model. The model takes text embeddings of entity names generated using the MiniLM-L6-v2 sentence transformer as input to predict knowledgeability scores for all the entities across the dataset. Further, we define an entity’s ``ignorance'' as one minus its predicted knowledgeability. Entities with the highest ignorance are preferred for fine‑tuning, and ties are broken first by choosing the entity with the lowest graph degree, to encourage topical diversity, and finally at random. We iteratively add entities and their associated triplets until 80\% of the budget is filled. In case an entity's full triplet set would overshoot the remaining slots, we randomly sample within that set to exactly meet the quota. The full Targeted training set thus comprises the initial 20\% query triples plus the 80\% ignorance‑weighted triplets.
    
    \item \textbf{Random Fine-Tuning:} For the Random Fine-Tuning, we retain the initial 20\% query set and additionally randomly sample the remaining 80\% of triplets from all unprobed triplets without replacement. This yields a direct random selection comparison to the targeted method.
\end{itemize}

\section{Results across different LLMs}\label{app-result}
\subsection{Llama 3.3 70B}
\begin{figure}[t!]
    \centering
    \includegraphics[width=1\linewidth]{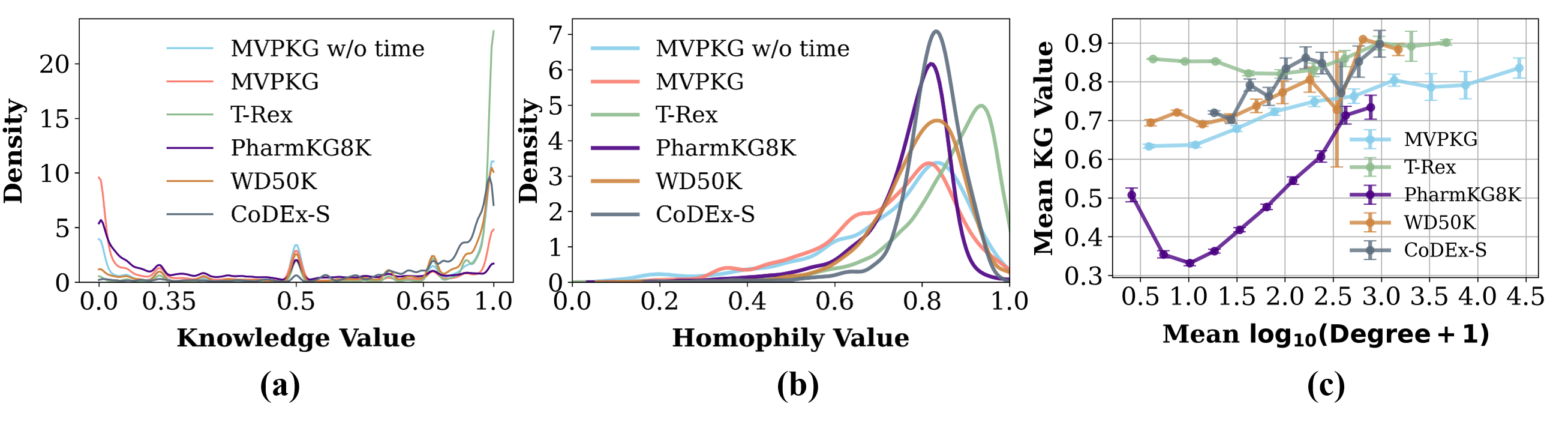}
    \vspace{-4ex}
    \caption{LLaMa \textbf{(a)}: Distribution of node knowledgeability for each dataset; \textbf{(b)}: Distribution of node homophily for each dataset; \textbf{(c)}: Node knowledgeability increases as node degree increases. }
    \label{fig-analysis-llama}
\end{figure}

See Figure~\ref{fig-analysis-llama} for an overview of model behavior.
\begin{itemize}[leftmargin=*]
    \item \textbf{Knowledgeability Distribution:} Similar to GPT3.5 results, Llama 3.3 70B has a trimodal pattern in the knowledgeability distribution, with domain-specific datasets having higher peaks at 0 while general datasets like T-Rex, which are extracted from Wikipedia, have higher peaks at 1. Peak at 0.5 is largely made of entities with degree 2 where one triplet is evaluated as true while the other one as false.
    \item \textbf{Homophily Distribution:} All datasets have homophily peak at 0.8 and above indicating that nodes and their neighbors tend to share similar knowledgeability scores. We observe overall a higher homophily on the general domain datasets than domain specific ones.
    \item \textbf{Degree vs Knowledgeability:} We observe that all datasets overall have a positive trend between the mean knowledgeability value and mean log degree. Biomedical dataset PharmKG8K has a higher upward trend, while MVPKG has a much shallower trend. This might be attributed to the T-Rex dataset's origin from Wikipedia entities which are well covered by pre-training corpora. 
\end{itemize}

\subsection{Deepseek V3}
\begin{figure}[t!]
    \centering
    \includegraphics[width=1\linewidth]{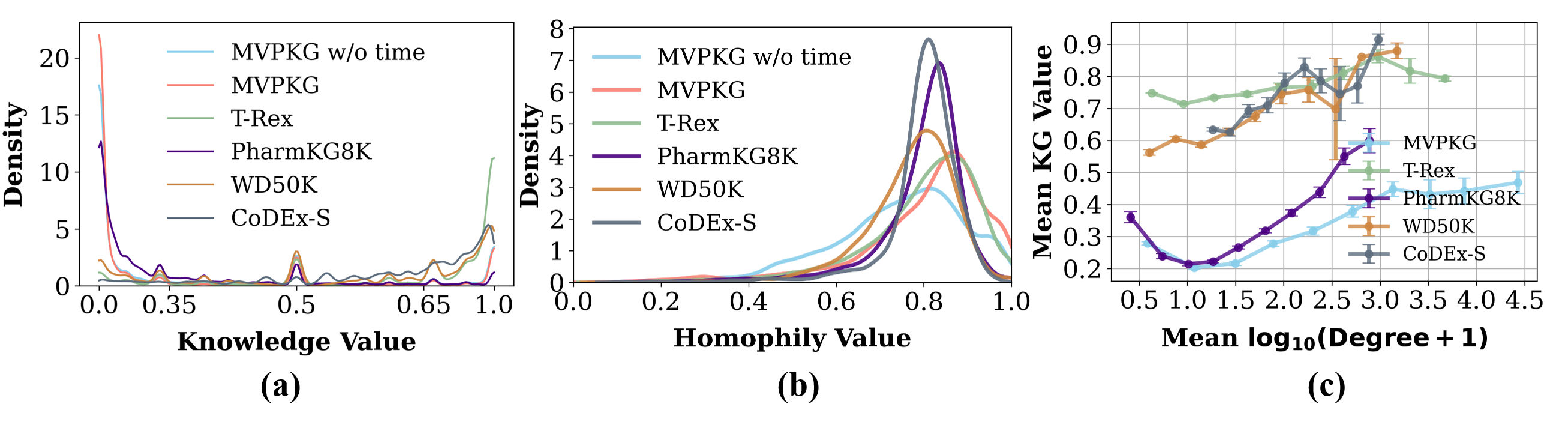}
    \vspace{-4ex}
    \caption{Deepseek \textbf{(a)}: Distribution of node knowledgeability for each dataset; \textbf{(b)}: Distribution of node homophily for each dataset; \textbf{(c)}: Node knowledgeability increases as node degree increases. }
    \label{fig-analysis-deepseek}
\end{figure}

See Figure~\ref{fig-analysis-deepseek} for an overview of model behavior.
\begin{itemize}[leftmargin=*]
    \item \textbf{Knowledgeability Distribution:} We observe a trimodel pattern with a relatively small peak at 0.5. Entities with a knowledge value of 0 are more common than those with a value of 1, especially in domain-specific datasets. For general datasets like T-Rex, WD50K, and CoDEx-S, a larger proportion of their entities are still recognized by Deepseek, resulting in a higher peak in the number of entities with full knowledgeability.
     \item \textbf{Homophily Distribution:} Homophily for entities across datasets has the highest density at around 0.8, indicating that entities and their neighbors tend to share similar knowledgeability scores. Here, no specific datasets appear to have a clear advantage over others. 
     \item \textbf{Degree vs Knowledgeability:} All datasets show a clear positive trend between the degree of entity and their Knowledgeability. T-Rex here has a slightly steeper trend than both GPT 3.5 and Llama 3.3 70 B.
\end{itemize}

\subsection{Gemini 2.5 Flash}
\begin{figure}[t!]
    \centering
    \includegraphics[width=1\linewidth]{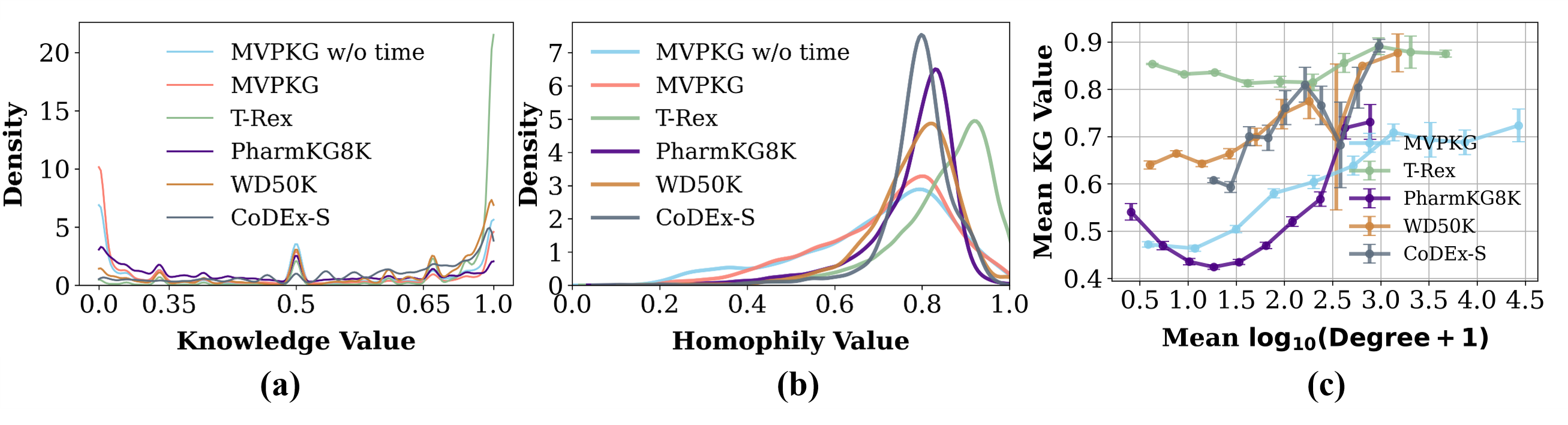}
    \vspace{-4ex}
    \caption{Gemini \textbf{(a)}: Distribution of node knowledgeability for each dataset; \textbf{(b)}: Distribution of node homophily for each dataset; \textbf{(c)}: Node knowledgeability increases as node degree increases. }
    \label{fig-analysis-gemini}
\end{figure}

See Figure~\ref{fig-analysis-gemini} for an overview of model behavior.
\begin{itemize}[leftmargin=*]
    \item \textbf{Knowledgeability Distribution:} The general-domain datasets continue to have a higher proportion of entities with a knowledgeability score of 1, resulting in a right-skewed distribution. In contrast, domain-specific datasets show a higher proportion of entities with a knowledgeability score of 0. A notable improvement of Gemini is that PharmKG8K has a more balanced distribution compared to the other models, like Llama 3.3 70B, GPT3.5, and Deepseek V3. This indicates that it has better knowledge about biomedical-related entities. Although there is still some left skew, it is significantly less pronounced.
    \item \textbf{Homophily Distribution:} Similar to other models, highest homophily density stays around 0.8, suggesting that nodes tend to have similar knowledgeability scores as their neighbors. T-Rex has a homophily to the furthest right, further indicating the nodes have very similar knowledge values to their neighbors.
    \item \textbf{Degree vs Knowledgeability:} A positive trend is observed across all datasets, with each showing an upward-sloping pattern. T-Rex, while following this trend, displays a relatively shallow slope, consistent with the behavior seen in other models, due to it being derived from Wikipedia. 
\end{itemize}

\subsection{GPT 4o}
\begin{figure}[t!]
    \centering
    \includegraphics[width=1\linewidth]{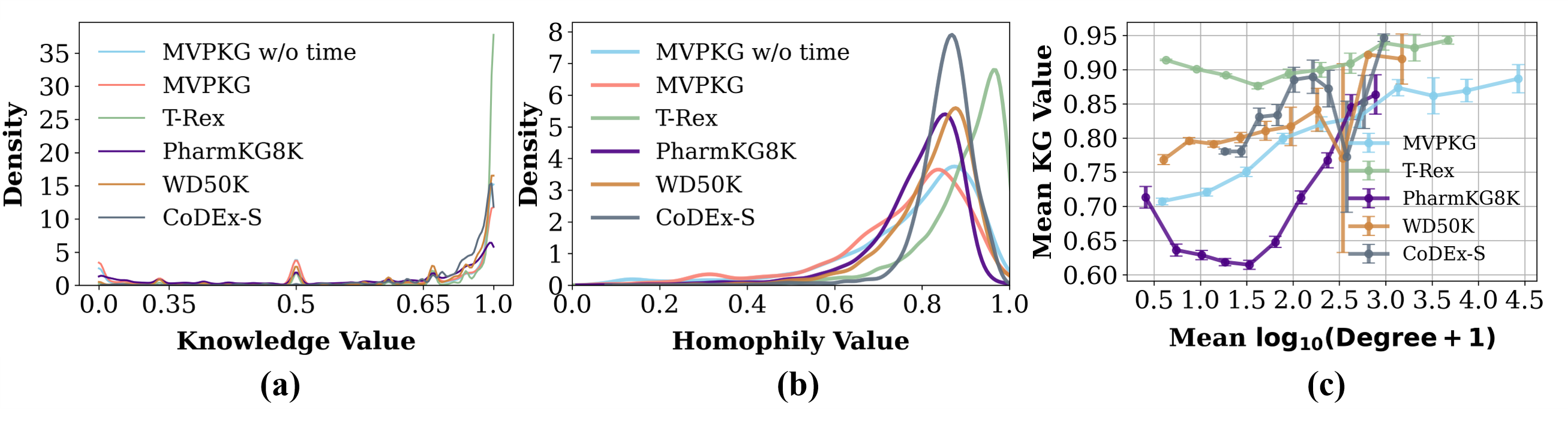}
    \vspace{-4ex}
    \caption{GPT4o \textbf{(a)}: Distribution of node knowledgeability for each dataset; \textbf{(b)}: Distribution of node homophily for each dataset; \textbf{(c)}: Node knowledgeability increases as node degree increases. }
    \label{fig-analysis-gpt4o}
\end{figure}

See Figure~\ref{fig-analysis-gpt4o} for an overview of model behavior.
\begin{itemize}[leftmargin=*]
    \item \textbf{Knowledgeability Distribution:} GPT-4o demonstrates a higher level of entity knowledgeability across all domains compared to other models. Even in domain-specific datasets like PharmKG8K, GPT-4o recognizes a larger proportion of entities than it does not. 
    \item \textbf{Homophily Distribution:} Here, datasets display a high level of homophily, with the highest density peaks being greater than 0.8. Following the pattern across the models, the T-Rex dataset presents the highest homophily among all the other datasets.
    \item \textbf{Degree vs Knowledgeability:} All datasets exhibit an upward trend, suggesting that as the degree associated with an entity increases, so does its knowledgeability. 
\end{itemize}


\section{KG vs Topology Analysis across models}
\begin{figure*}[htbp!]
    \centering
    \includegraphics[width=1\linewidth]{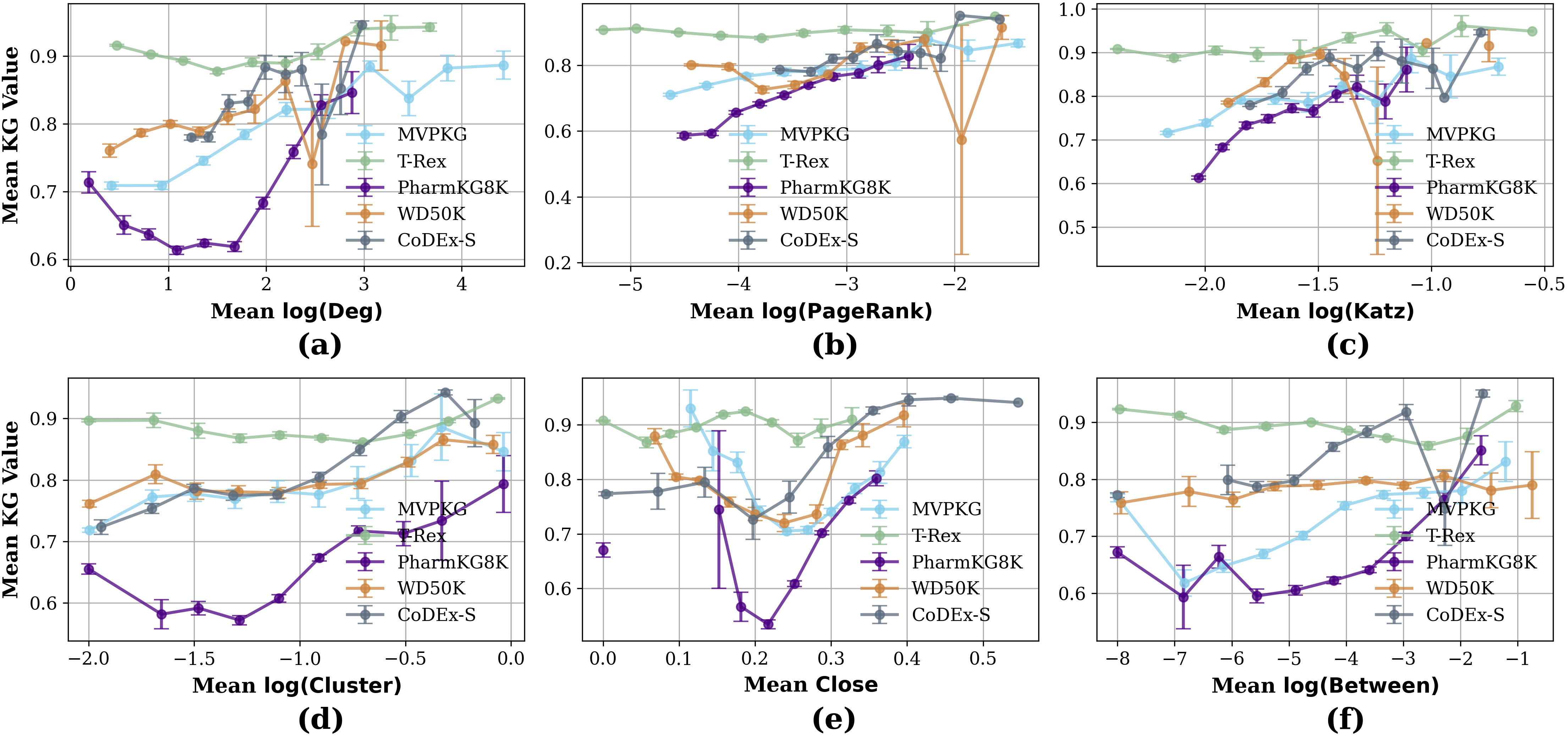}
    \vspace{-5ex}
    \caption{GPT4o - Relationship between Mean Knowledgeability and \textbf{(a)}: Degree Centrality; \textbf{(b)}: PageRank Centrality; \textbf{(c)}: Katz Centrality; \textbf{(d)}: Cluster Centrality; \textbf{(e)}: Closeness Centrality; \textbf{(f)}: Betweeness Centrality.}
    \label{fig-topo-analysis-gpt4o}
    \vspace{-2ex}
\end{figure*}

\begin{figure*}[htbp!]
    \centering
    \includegraphics[width=1\linewidth]{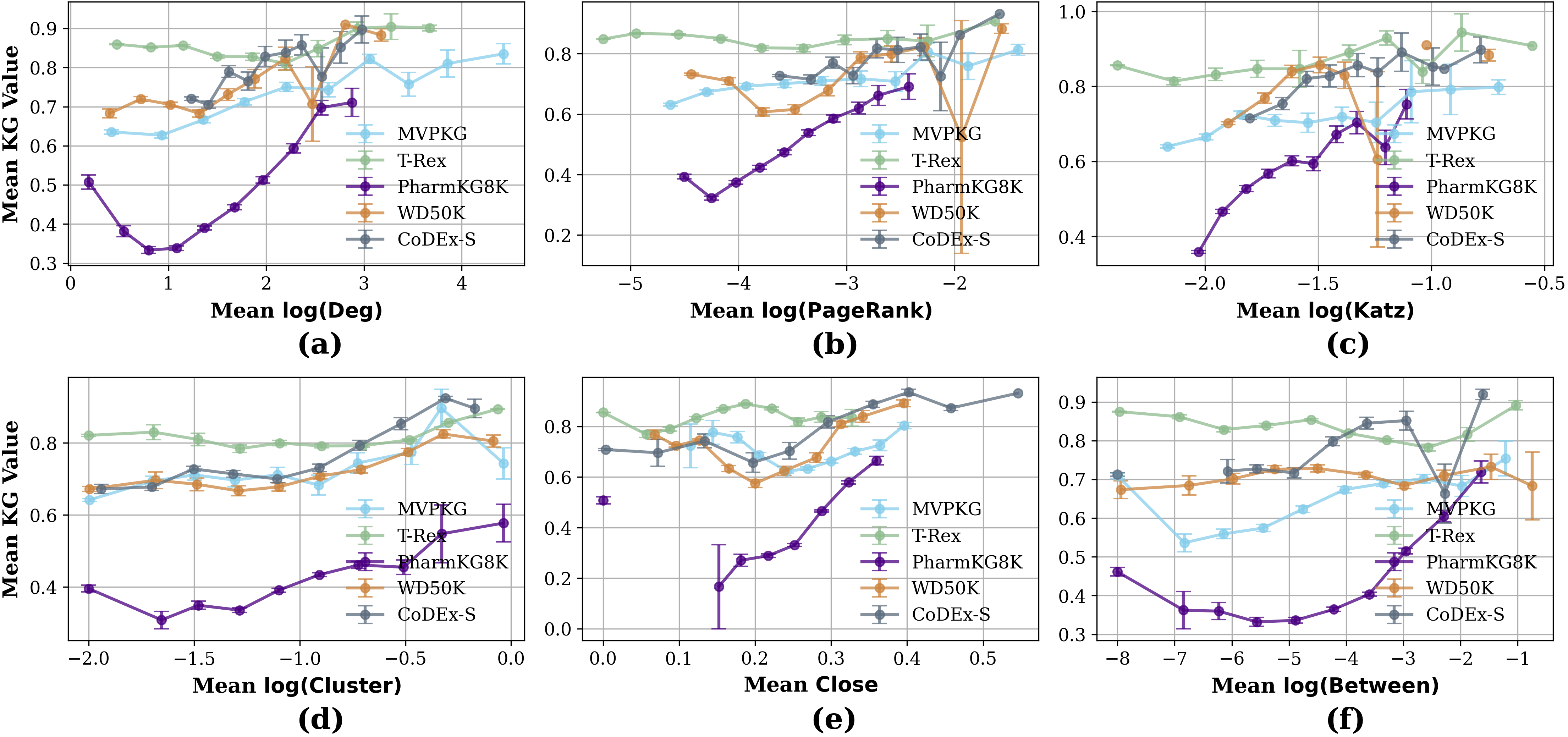}
    \vspace{-5ex}
    \caption{LLaMa - Relationship between Mean Knowledgeability and \textbf{(a)}: Degree Centrality; \textbf{(b)}: PageRank Centrality; \textbf{(c)}: Katz Centrality; \textbf{(d)}: Cluster Centrality; \textbf{(e)}: Closeness Centrality; \textbf{(f)}: Betweeness Centrality.}
    \label{fig-topo-analysis-llama}
    \vspace{-2ex}
\end{figure*}

\begin{figure*}[htbp!]
    \centering
    \includegraphics[width=1\linewidth]{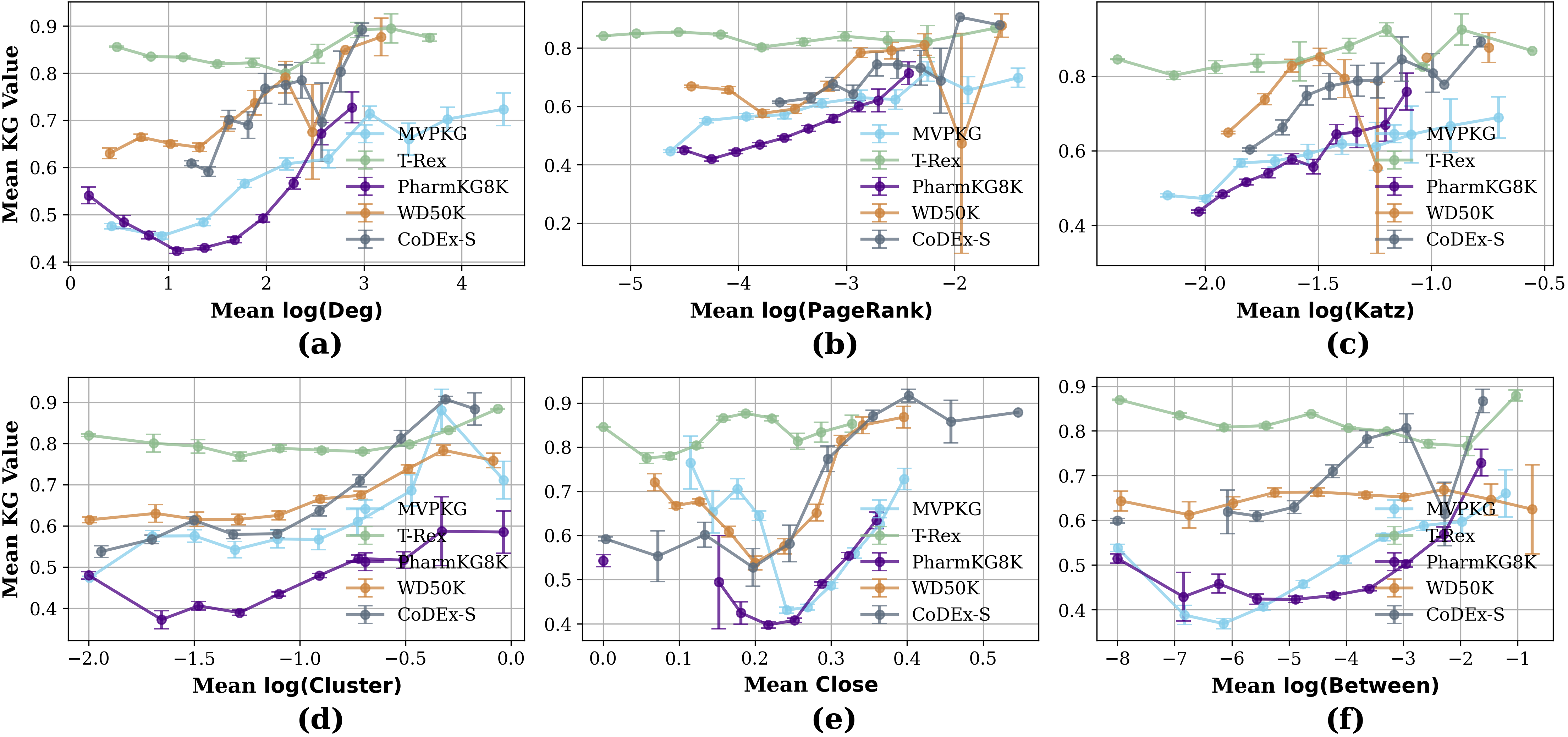}
    \vspace{-5ex}
    \caption{Gemini - Relationship between Mean Knowledgeability and \textbf{(a)}: Degree Centrality; \textbf{(b)}: PageRank Centrality; \textbf{(c)}: Katz Centrality; \textbf{(d)}: Cluster Centrality; \textbf{(e)}: Closeness Centrality; \textbf{(f)}: Betweeness Centrality.}
    \label{fig-topo-analysis-gemini}
    \vspace{-2ex}
\end{figure*}

\begin{figure*}[t!]
    \centering
    \includegraphics[width=1\linewidth]{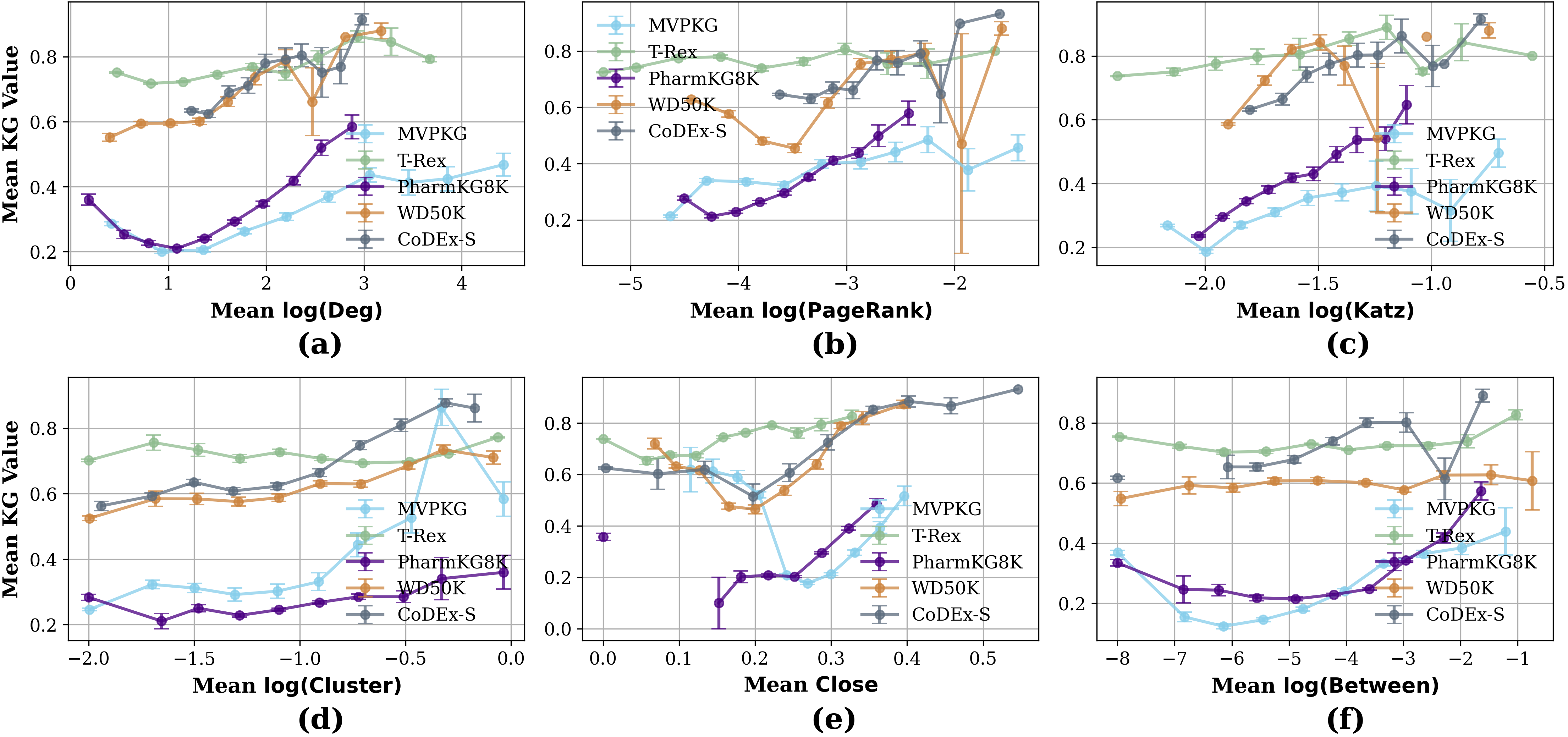}
    \vspace{-4ex}
    \caption{Deepseek - Relationship between Mean Knowledgeability and \textbf{(a)}: Degree Centrality; \textbf{(b)}: PageRank Centrality; \textbf{(c)}: Katz Centrality; \textbf{(d)}: Cluster Centrality; \textbf{(e)}: Closeness Centrality; \textbf{(f)}: Betweeness Centrality.}
    \label{fig-topo-analysis-deepseek}
    \vspace{-1ex}
\end{figure*}

\begin{figure*}[t!]
     \centering
     \includegraphics[width=1\linewidth]{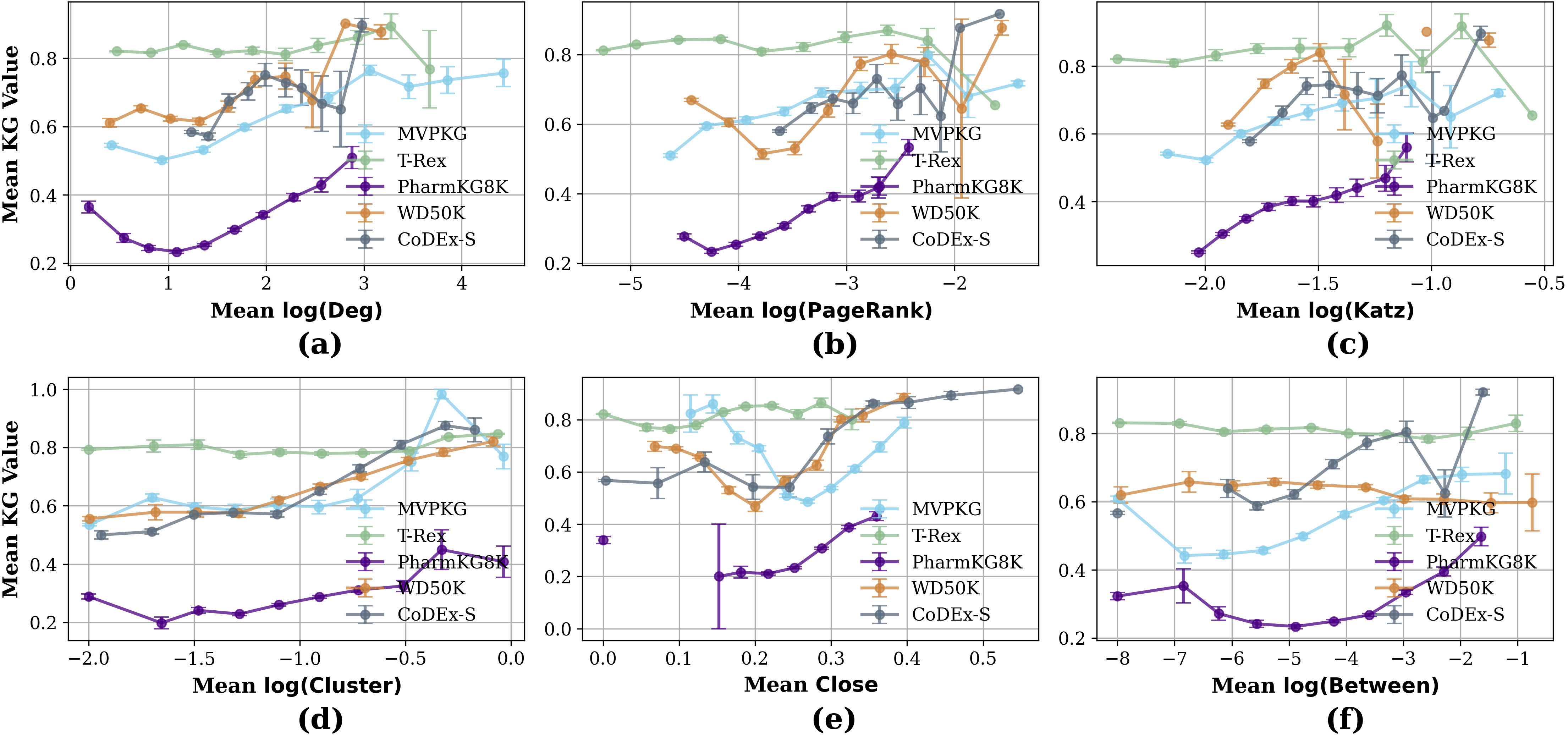}
     \vspace{-4ex}
     \caption{GPT3.5 - Relationship between Mean Knowledgeability and \textbf{(a)}: Degree Centrality; \textbf{(b)}: PageRank Centrality; \textbf{(c)}: Katz Centrality; \textbf{(d)}: Cluster Centrality; \textbf{(e)}: Closeness Centrality; \textbf{(f)}: Betweeness Centrality.}
     \label{fig-analysis-3.5}
     \vspace{-1ex}
 \end{figure*}

For each node, we calculate its corresponding graph structural properties and group them based on these properties. For each group, we further calculate the average knowledge and visualize its relation with structural properties.
See Figure~\ref{fig-topo-analysis-gpt4o} for GPT-4o; Figure~\ref{fig-topo-analysis-llama} for Llama 3.3 70B; Figure~\ref{fig-topo-analysis-gemini} for Gemini 2.5; and Figure~\ref{fig-topo-analysis-deepseek} for DeepSeek V3 for an overview; Figure~\ref{fig-analysis-3.5} for GPT3.5.

\begin{itemize}[leftmargin=*]
    \item \textbf{Degree Centrality:} We observe a general positive upward trend among the mean knowledge value and degree centrality across the models. A high degree node would appear in many facts and would appear in large amount of training corpus. Therefore, if sample those corpus for training the model, that entity would show up at more places and the model would get more examples of the entity, and thus learning about it better.
    
    \item \textbf{PageRank Centrality:} Here, across the models we observe a positive trend. WD50K displays a large variance towards the top. Since, the bins there contain few entities, variance is presented as large in case of any outlier.
    
    \item \textbf{Katz Centrality:} We observe a positive trend among 4 out of 5 datasets. WD50K creates a upside down u shape slope with some outliers, along with high variance. This can potentially be attributed to a few entities in the last few bins presenting an increased variance and unexpected behavior.
    
    \item \textbf{Cluster Centrality:} Across the models we see a positive trend between the mean knowledgeability value and the cluster centrality. This can potentially be caused by the fact that a higher clustering would mean that entity is part of a dense group and would be mentioned over and over whenever the context of that group comes up. However, the rate is less pronounced in some than in others. For example, GPT 4o has a stronger relationship trend than Deepseek V3. T-Rex, for all the models has a very slight but positive trend, mostly staying relatively flat. 
    
    \item \textbf{Closeness Centrality:} Here, the results vary the most. For GPT 4o, almost all datasets have a U-shape, indicating that both peripheral and central nodes get higher knowledgeability values. In contrast, the Llama model has a relatively minor U-shape effect, with some datasets broadly staying flat, and for example, PharmKG8K showing an upward trend.
    
    \item \textbf{Between Centrality:} Here, datasets with a general domain like WD50K and T-Rex stay relatively flat, whereas domain-specific datasets, such as PharmKG8K, display a strong positive relation, indicating that entities that serve as hubs or bridges tend to have a higher knowledgeability score than nodes on the periphery. 
\end{itemize}

\newpage
\section{Temporal LLM-based Triplet Evaluation Prompt}\label{app-temporal-prompt}
\begin{tcolorbox}[
  colback=blue!10!white,
  colframe=blue!80!black,
  title=Prompt 2: LLM‐based Triplet Evaluation (Temporal Variation),
  boxsep=0.75mm, left=0.75mm, right=0.75mm, top=0.75mm, bottom=0.75mm,
  float=htbp!, floatplacement=htbp!,
]
\scriptsize
\textbf{System Message:} Evaluate the statement below; reply only \texttt{True} or \texttt{False}.\\[2pt]
\textbf{Given:} 
Triplet $\mathcal{T}=(\mathit{sub},\,\mathit{rel},\,\mathit{obj})$, 
Date $D$.\\[2pt]
\textbf{Relational Template Map:} 
$T:\,\mathit{rel}\mapsto\text{“\{\textit{sub}\} … \{\textit{obj}\}”}$.\\[2pt]
\textbf{Procedure:}
\begin{enumerate}[nosep,left=6pt]
  \item Retrieve template $t = T(\mathit{rel})$.
  \item Instantiate base statement  
    $S_0 = t[\{\mathit{sub}\}\!\to\!\mathit{sub},\,\{\mathit{obj}\}\!\to\!\mathit{obj}]$.
  \item Append date: 
    $S = S_0\ \text{on}\ D$.
  \item Send \textbf{System Msg} + \textbf{User Msg:} $S$ to LLM.
  \item Return “True” or “False.”
\end{enumerate}
\label{box:prompt_template2}
\end{tcolorbox}



\end{document}